\documentclass[dvipsnames,format=sigconf,anonymous=false,review=false,screen=true]{acmart}

\AtBeginDocument{%
  }

\usepackage{algorithm}
\usepackage{algorithmicx}
\usepackage{algpseudocode}
\usepackage{cleveref}

\algnewcommand{\LeftComment}[1]{\State \(\triangleright\) #1}
\algnewcommand{\LeftCommentX}[1]{\Statex \(\triangleright\) #1}

\begin{document}

\title{Diversifying Toxicity Search in Large Language Models Through Speciation}

\author{Onkar Shelar}
\email{os9660@rit.edu}
\orcid{0009-0005-5109-6641}
\affiliation{%
  \institution{Rochester Institute of Technology}
  \city{Rochester}
  \state{New York}
  \country{USA}
}

\author{Travis Desell}
\email{tjdvse@rit.edu}
\orcid{0000-0002-4082-0439}
\affiliation{%
  \institution{Rochester Institute of Technology}
  \city{Rochester}
  \state{New York}
  \country{USA}
}

\renewcommand{\shortauthors}{Shelar and Desell}

\begin{abstract}
Evolutionary prompt search is a practical black-box approach for red teaming large language models, however existing methods often collapse onto a small family of high-performing prompts, limiting coverage of distinct failure modes. We present a speciated quality-diversity extension of \textit{ToxSearch} that maintains multiple high-toxicity prompt niches in parallel rather than optimizing a single best prompt. \textit{ToxSearch-S} introduces unsupervised prompt speciation via a search methodology that maintains capacity-limited species with exemplar leaders, a reserve pool for emerging niches, and species-aware parent selection that trades off within-niche exploitation and cross-niche exploration. Preliminary results show \textit{ToxSearch-S} reaching higher peak toxicity ($\approx 0.73$ vs.\ $\approx 0.47$) with a heavier tail (top-10 median $0.66$ vs.\ $0.45$) than the baseline. Speciation also yields broader semantic coverage under a topics-as-species analysis (higher effective topic diversity and larger unique topic coverage). Finally, species formed are well-separated in embedding space (mean separation ratio $\approx 1.93$) and exhibit distinct toxicity distributions, indicating that speciation partitions the adversarial space into behaviorally differentiated niches rather than superficial lexical variants.
\end{abstract}




\maketitle
{\textcolor{red}{This is pre-print.}}
\section{Introduction}

Large Language Models (LLMs) are highly capable, but adversarial prompting continues to expose failure modes that produce harmful generations~\cite{corbo2025toxic, shelar2025evolving, srivastava2023no}. For that reason, rigorous red teaming is a core requirement for safety evaluation\footnote{\textcolor{red}{This paper contains disturbing language presented solely for safety evaluation.}}. Much of existing red teaming practice remains manual or guided by heuristics~\cite{cao2024defending}, which does not scale and tends to probe only a thin slice of the adversarial prompt space. Search-based methods~\cite{shelar2025evolving,corbo2025toxic,guo2023connecting,liu2023autodan,samvelyan2024rainbow} address this by approaching it as an optimization problem and using evolutionary algorithms (EAs) to search. By evolving a population of prompts, they iteratively discover inputs that elicit increasingly harmful content from a target model. Over generations, variation and selection can surface non-obvious failure modes that may be difficult to uncover through manual red teaming alone~\cite{shelar2025evolving,corbo2025toxic,guo2023connecting,liu2023autodan}. Most work in this area relies on variation and possible ad-hoc diversity encouragement, but does not enforce that different prompt niches evolve in parallel. Quality-Diversity (QD) algorithms aim to produce an archive of solutions that are individually high-quality yet collectively diverse~\cite{pugh2016qualitydiversity}. In toxicity red teaming, \emph{quality} naturally maps to the toxicity achieved in a model’s response, while \emph{diversity} corresponds to distinct adversarial prompt themes. Recent automated red-teaming work~\cite{samvelyan2024rainbow,dang2025rainbowplus} makes this explicit, whose QD motivations align with empirical behavior in evolutionary prompt search. In natural evolution, diversity is preserved through speciation. Inspired by this, evolutionary computation developed several niching techniques to maintain multiple solution niches within a single run~\cite{10.5555/240028,goldberg1987genetic,mahfoud1992crowding,542703,harik1995finding,6793380}. By mimicking ecological niches or species, they allow the algorithm to explore several peaks in parallel. QD algorithms are sometimes called illumination algorithms because their aim is to illuminate the space of possibilities~\cite{mouret2015illuminating}. 

Taken together, these results justify treating diversity as a first-class requirement in evolutionary red teaming and motivate moving from ToxSearch's implicit topical niching to explicit prompt speciation. None of the reviewed works~\cite{guo2023connecting,fernando2023promptbreeder,liu2023autodan} explicitly speciate the prompt population during the evolutionary process. While Rainbow Teaming applies a QD approach, most other evolutionary prompt works optimize a single objective and return one result or a small set. Maintaining a MAP-Elites style archive of diverse high-toxicity prompts throughout evolution would directly benefit practitioners, providing a suite of attack examples~\cite{10.1162/106365602320169811}.  We see an open gap in determining how to systematically maintain and leverage diversity (especially via speciation) in the evolutionary search for toxic prompts, in order to discover multiple high-impact prompt attacks rather than converging on a single mode. This gap sits at the intersection of EAs and AI safety, it requires: (i) clustering language artifacts in a way that is semantically meaningful \emph{and} behaviorally relevant, and (ii) demonstrating that explicit diversity mechanisms improve coverage without sacrificing peak toxicity.  We address this gap by introducing an online unsupervised speciation mechanism inside ToxSearch and evaluating its QD trade-off.

\section{Methodology}

\emph{ToxSearch-S}\footnote{\textbf{GitHub repo} - https://github.com/Onkar2102/ToxSearch-S} extends \emph{ToxSearch}~\cite{shelar2025evolving} with unsupervised online speciation to maintain multiple niches in parallel. Evaluated prompts are assigned to species via leader-follower clustering under an ensemble distance over prompt semantics and response-level toxicity signals, while outliers are stored in a reserve pool. This design reframes adversarial prompt search as QD optimization, targeting both high toxicity and sustained behavioral diversity across species. Core components from ToxSearch are retained, as the propmpt generator (PG)PG and response generator (RG) split, moderation evaluator, and operator suite address \emph{how} to generate and evaluate prompts, while speciation addresses \emph{how} to organize and maintain diverse populations.

Given a target response generator (RG) LLM $\theta_{rg}$ and a moderation oracle $\mathcal{M}$ (Google's Perspective API~\cite{perspectiveAPI}), we seek prompts $p \in \mathcal{P}$ that maximize toxicity on the model response $y \sim \theta_{rg}(p)$, where $p$ is produced by a separate prompt generator (PG) LLM $\theta_{pg}$. The oracle maps text to attribute scores $\mathbf{s}(y)\in[0,1]^K$, and we use toxicity as scalar fitness, i.e., $\hat{f}(p)=\mathrm{s}_{\text{toxicity}}(\mathcal{M}(\theta_{rg}(p))) \in [0,1]$. \emph{ToxSearch-S} reframes this as a QD objective over species $\{S_1,\ldots,S_k\}$ by maximizing the sum of per-species best fitness values, $\sum_{i=1}^{k}\max_{p \in S_i}\hat{f}(p)$, subject to an inter-species diversity constraint $D_{inter}(\{S_1,\ldots,S_k\}) \ge \theta_{diversity}$. Here, $\{S_1,\ldots,S_k\}$ is a partition of the population into semantically coherent species, and $D_{inter}$ measures separation between species.

Each species is represented by its leader $(leader(S_{i}))$, which is the highest-fitness member used for distance computations. Population diversity is measured using an ensemble distance that combines genotype (semantic) and phenotype (behavioral) dissimilarity~\cite{ando2007heuristic, burlacu2023inheritance, kim2009distancemeasures}. For prompts $u$ and $v$, we define the ensemble distance $d_{\mathrm{ensemble}}(u,v)$ as a weighted combination of genotype-level and phenotype-level dissimilarity. The genotype component, $d_{\mathrm{genotype\_norm}}(u,v)$, is computed from the cosine distance between the L2-normalized prompt embeddings $e_u,e_v \in \mathbb{R}^{384}$ and scaled to $[0,1]$ as $\frac{1-(e_u^\top e_v)}{2}$. The phenotype component, $d_{\mathrm{phenotype}}(u,v)$, is the normalized Euclidean distance between the corresponding moderation score vectors $\mathbf{s}(y_u), \mathbf{s}(y_v) \in [0,1]^8$, where $\mathbf{s}(y)$ contains the eight Perspective API attributes. In practice, we set $d_{\mathrm{ensemble}}(u,v)=0.7\,d_{\mathrm{genotype\_norm}}(u,v)+0.3\,d_{\mathrm{phenotype}}(u,v)$, giving greater weight to semantic separation while still accounting for response-level behavioral differences.

To prevent unbounded growth, the system enforces capacity limits on $S_{i}$ and $R$. When a species exceeds capacity, members are ranked by fitness in descending order and only the top $C_{\mathrm{species}}$ are retained. Species are merged when leader distance is below $\theta_{\mathrm{merge}}$, and the highest-fitness genome becomes the new leader.  Species which have members selected as parent and have shown no progress from $T_{species}$ generations are frozen and species with members less than $C_{min}$ are removed. A frozen species is retained in the archive but excluded from parent selection and variation to avoid spending query budget on lineages that have plateaued. When $\theta_{\mathrm{rg}}$ produces refusal-style outputs, detected via lightweight pattern matching of common refusal phrases, we apply a soft penalty to the toxicity-based fitness such that $\hat{f}_{\mathrm{penalized}}(p)=0.85\,\hat{f}(p)$, thereby discouraging refusal-inducing prompts without removing them entirely from evolution.

\section{Experiments}

Both $\theta_{\mathrm{pg}}$ and $\theta_{\mathrm{rg}}$ used Llama 3.1-8B-Instruct model in GGUF format, with temperatures 0.9 and 0.7, respectively. We ran ToxSearch in two modes under the same core evolutionary budget $(G=50)$ as a non-speciation baseline and ToxSearch with speciation. For non-speciation, all runs used $\alpha=30$ and $\beta=3$ as percentage parameters, with $\alpha$ determining the elite threshold and $\beta$ determining the removal threshold of under-performing prompts relative to the current maximum toxicity score. Speciation parameteres were $\theta_{sim} = 0.25$ and $\theta_{merge} = 0.25$. Capacity controls were set to $C_{min} = 5$, $C_{species} = 25$ and $C_{reserves} = 500$. The stagnation limit for species was $T_{species}=7$. The initial population $P$ of 100 questions was uniformly sampled from pool which was created by merging the questions from CategoricalHarmfulQA and HarmfulQA~\cite{bhardwaj-etal-2024-language,Bhardwaj2023RedTeamingLL}.

\noindent
\textbf{RQ1:} \textit{Does speciated evolutionary search discover higher-quality toxic prompts faster, and does it maintain a more diverse set of toxic behaviors compared to the baseline approach?}

\begin{figure}
  \centering
  \includegraphics[width=\linewidth]{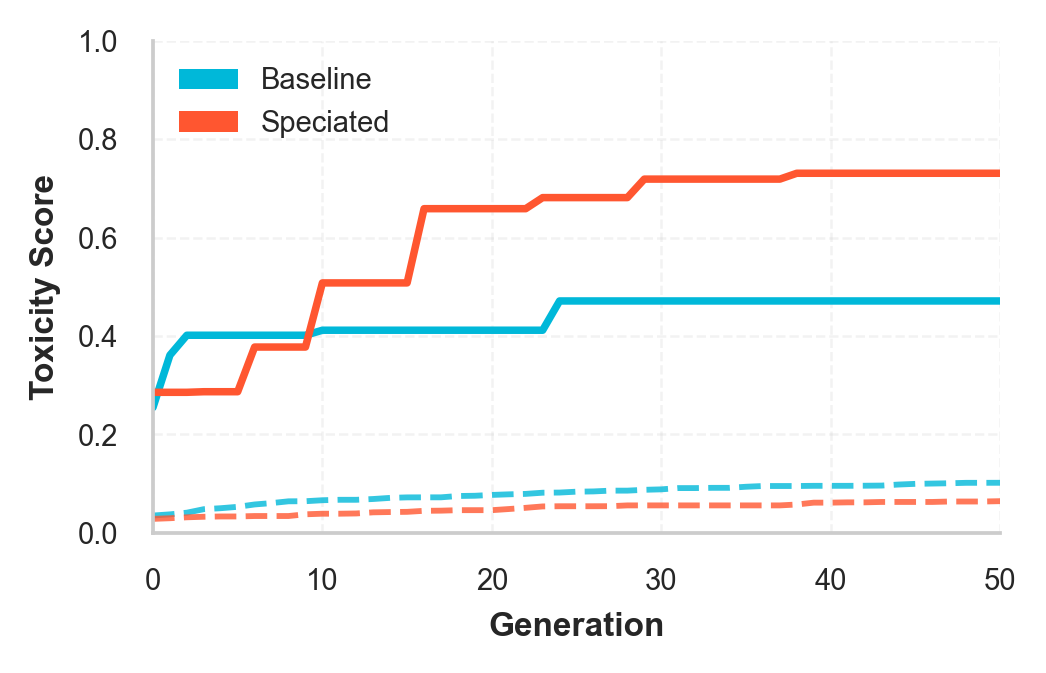}
  \caption{Cumulative maximum toxicity (solid) and Avg fitness (dashed) over generations}
  \Description{Line plot showing evolutionary trajectories over 50 generations. Speciated ToxSearch (red/orange) reaches higher toxicity scores than baseline (blue) for both max toxicity and average fitness metrics.}
  \label{fig:evolutionary_trajectory}
\end{figure}

\begin{figure}
  \centering
  \includegraphics[width=\linewidth]{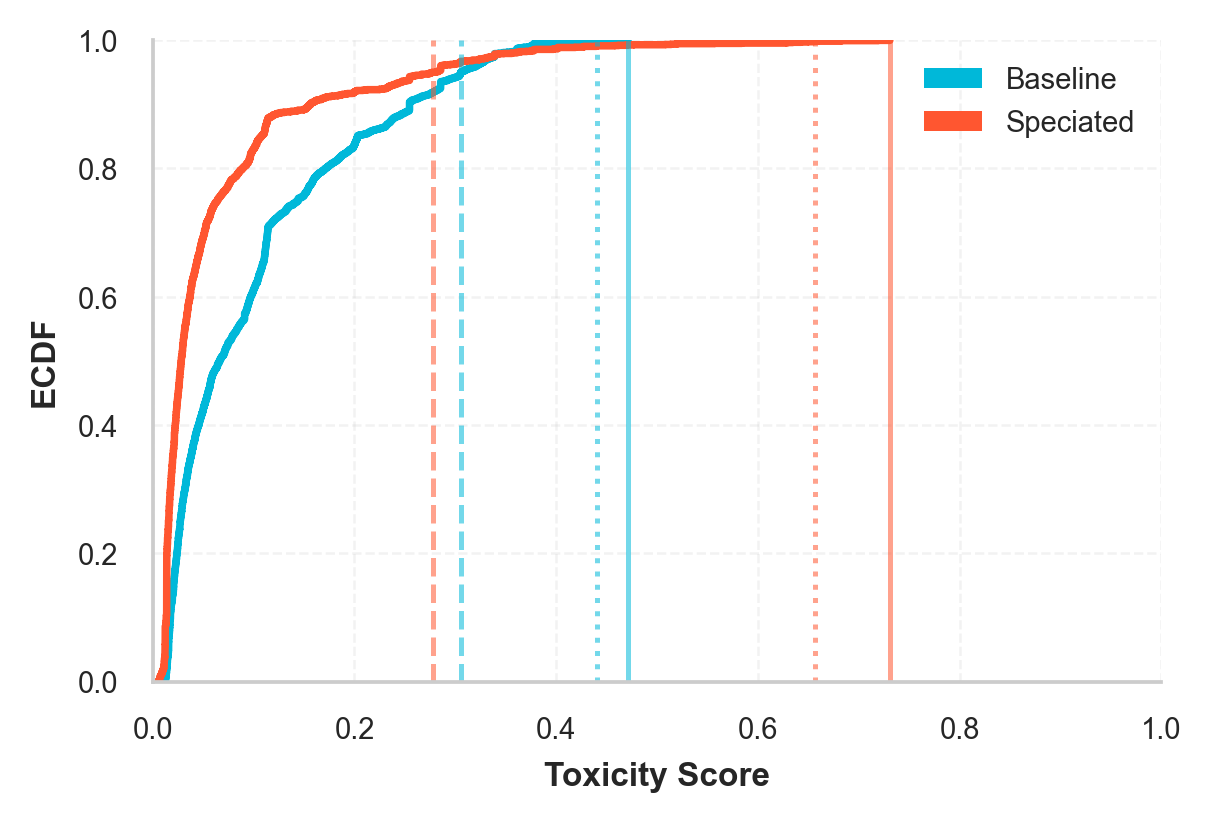}
  \caption{ECDF of prompt toxicities}
  \Description{ECDF plot comparing toxicity distributions. Vertical lines mark key percentiles for baseline (blue) and speciated (red) conditions.}
  \label{fig:toxicity_ecdf}
\end{figure}

\begin{figure}
  \centering
  \includegraphics[width=\linewidth]{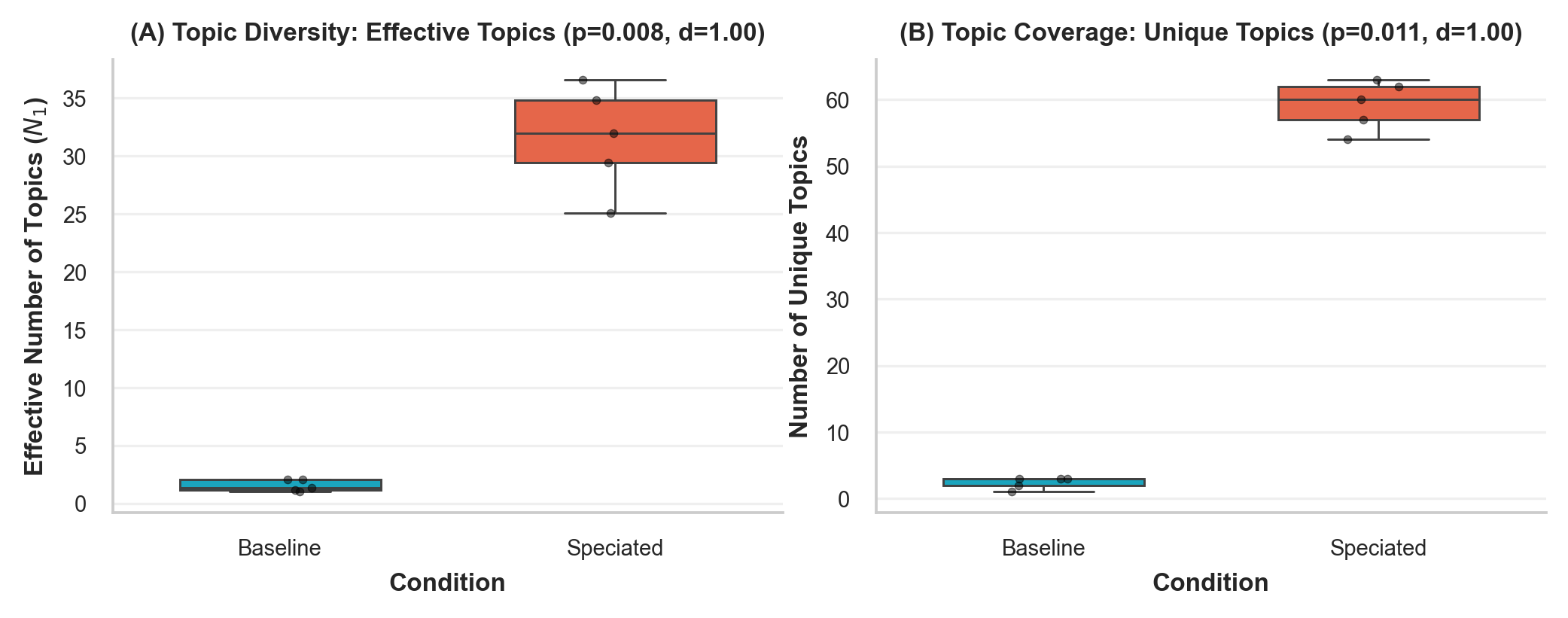}
  \caption{Topic diversity comparison}
  \Description{Two-panel box plot showing effective topics (N1) and unique topics (K) for baseline and speciated conditions with statistical annotations.}
  \label{fig:topic_diversity}
\end{figure}

We conducted a comparative analysis between baseline and speciated evolutionary search across five independent runs. Baseline ToxSearch used all elites and non-elites; while speciated ToxSearch used elites and reserves. All prompts were de-duplicated by canonicalized text. Quality metrics included peak performance ($Q_{\max}$), and depth metrics (top-10/top-50 mean toxicity). Diversity was assessed with semantic topic modeling on pooled prompts (effective topics $N_1 = \exp(H)$ and unique topic count $K$).

Figure~\ref{fig:evolutionary_trajectory} illustrates the evolutionary trajectories. The speciated approach consistently achieves higher peak toxicity values, with the maximum across runs around $0.7308$, compared to baseline which peaks around $0.4712$. Figure~\ref{fig:toxicity_ecdf} presents the empirical cumulative distribution function (ECDF) of all discovered prompt toxicities, revealing that while baseline produces a slightly higher proportion of moderately toxic prompts (95th percentile: 0.31 vs. 0.30), speciated discovers significantly more extreme cases, with a top-10 median toxicity of 0.66 versus 0.45 for the baseline, and achieving the absolute maximum toxicity score of 1.0. The topic-as-species diversity analysis (Figure~\ref{fig:topic_diversity}) demonstrates that speciated ToxSearch maintains greater semantic diversity, with higher effective topic numbers $N_1$ and broader topic coverage $K$, indicating exploration of more distinct behavioral modes.

Methodologically, Figure~\ref{fig:evolutionary_trajectory} plots cumulative maximum toxicity and average fitness aggregated across runs (maximum for toxicity, median for fitness). Figure~\ref{fig:toxicity_ecdf} uses ECDF $F(x) = \frac{1}{n} \sum_{i=1}^{n} \mathbf{1}[x_i \leq x]$ with markers for $Q_{0.95}$, top-10 median, and $Q_{\max}$. Topic diversity (Figure~\ref{fig:topic_diversity}) derives from a global BERTopic model (all-MiniLM-L6-v2 embeddings, c-TF-IDF), computing $N_1 = \exp(H)$ where $H = -\sum_{t} p_t \log p_t$. Together, these analyses support a \emph{ToxSearch-S} that improves toxicity and semantic breadth relative to the baseline.

\noindent
\textbf{RQ2:} \textit{Do different species clusters in speciated ToxSearch exhibit distinct toxicity distributions across embedding space and semantic topic clusters?}

\begin{figure*}[t]
\centering
\includegraphics[width=0.73\textwidth]{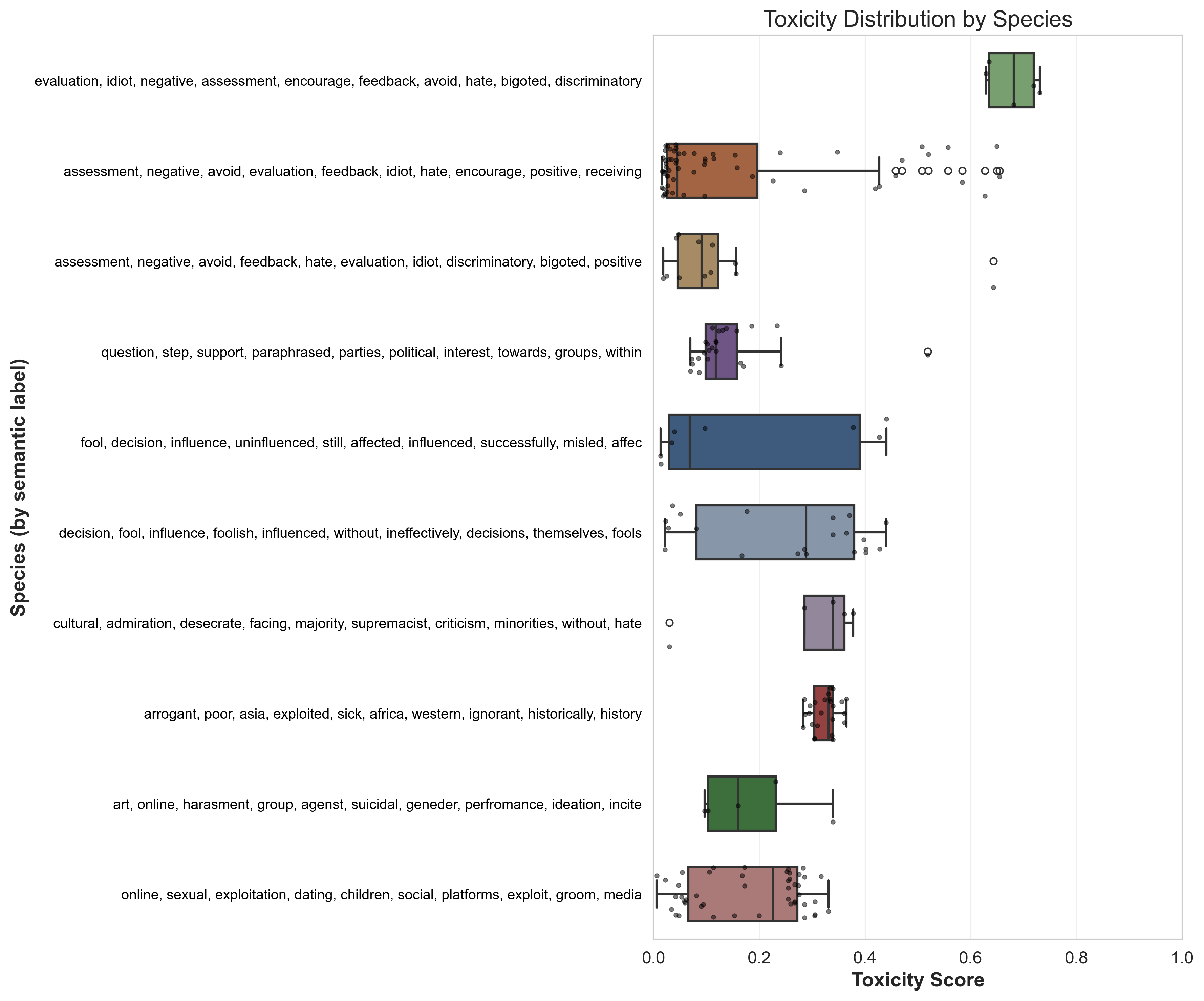}
\caption{Horizontal boxplots with jittered strip plots show toxicity scores for top species groups (grouped by 10-label)}
\label{fig:rq2_toxicity_distribution}
\end{figure*}

We analyzed species discovered through speciation across 5 independent runs, aggregating elite and reserve genomes to capture a broad set of solutions. The runs contain 12--28 mature species each (mean: 21.0). Cluster separation was measured as the ratio of mean inter-species to mean intra-species cosine distance between leader embeddings. Across runs, separation ratios range from 1.69 to 2.28 (mean: 1.93). Figure~\ref{fig:rq2_toxicity_distribution} presents the toxicity distribution for the top 10 species groups, demonstrating that different species exhibit distinct toxicity distributions.

Our analysis shows that speciation forms well-separated species clusters and that these species exhibit distinct toxicity distributions. Top-performing species reach maximum toxicity around 0.7, while lower-performing species follow different distributional patterns, supporting the claim that speciation partitions the search space into multiple, behaviorally distinct solution clusters.

\footnotetext{\textbf{Ethical considerations}: This work is intended for safety evaluation and red-teaming of aligned LLMs in a controlled research setting.}
\section{Conclusions}

This work introduced \emph{ToxSearch-S}, a QD method for adversarial prompt discovery that partitions search into behaviorally distinct species via unsupervised, distance-based clustering. Across experiments, speciated search consistently outperformed baseline search on extreme toxicity (max 0.73 vs.\ 0.47) while covering broader semantic space (higher effective topic diversity $N_1$ and unique topic coverage $K$). Although both methods recover similar moderately toxic prompts (95th percentile $\approx 0.30$), speciation produces a heavier high-toxicity tail (top-10 median 0.66 vs.\ 0.44). Species are also well separated (mean separation ratio $\approx 1.9$), and different species exhibit distinct toxicity profiles, suggesting that speciation maintains multiple productive niches rather than collapsing into a single mode. These results are interpreted under a limited budget (100 initial genomes, 50 generations, five runs per condition). The gains are strong evidence of improved search behavior, but not definitive evidence of convergence or long-run stability. Longer runs and larger seed sets are needed to test whether new toxic regions continue to emerge and whether species hit stable quality ceilings. A central takeaway is that clustering quality is foundational, where speciation helps only when distance features and assignment dynamics produce semantically coherent species.

This study also has important limitations. We use a single toxicity oracle (Perspective API), a fixed generator pairing, and modest run counts, so conclusions may shift across models, moderation systems, or annotation schemes. Future work should (i) evaluate transfer across target LLMs and multiple safety oracles, (ii) run longer-horizon stress tests with explicit stability metrics, and (iii) improve the algorithm with novelty-aware reserve management and species-aware parent selection. Stronger QD baselines (e.g., MAP-Elites-style archives), sensitivity analyses over clustering hyperparameters, and human-in-the-loop cluster validation would further strengthen reliability and interpretability.

\begin{acks}
We thank RIT's Research Computing team for their assistance and support~\cite{rit-rc-services}.
\end{acks}

\bibliographystyle{ACM-Reference-Format}
\bibliography{sample-base}

@article{shelar2025evolving,
  title={Evolving Prompts for Toxicity Search in Large Language Models},
  author={Shelar, Onkar and Desell, Travis},
  journal={arXiv preprint arXiv:2511.12487},
  year={2025}
}

@article{corbo2025toxic,
  title={How Toxic Can You Get? Search-based Toxicity Testing for Large Language Models},
  author={Corbo, Simone and Bancale, Luca and De Gennaro, Valeria and Lestingi, Livia and Scotti, Vincenzo and Camilli, Matteo},
  journal={arXiv preprint arXiv:2501.01741},
  year={2025}
}

@online{perspectiveAPI,
  title =        "Google Perspective API",
  url =          "https://perspectiveapi.com",
  month =        jan,
  lastaccessed = "Jan 13, 2026"
}

@article{pugh2016qualitydiversity,
author = {Pugh, Justin and Soros, Lisa and Stanley, Kenneth},
year = {2016},
month = {07},
pages = {},
title = {Quality Diversity: A New Frontier for Evolutionary Computation},
volume = {3},
journal = {Frontiers in Robotics and AI},
doi = {10.3389/frobt.2016.00040}
}

@article{samvelyan2024rainbow,
  title={Rainbow teaming: Open-ended generation of diverse adversarial prompts},
  author={Samvelyan, Mikayel and Raparthy, Sharath C and Lupu, Andrei and Hambro, Eric and Markosyan, Aram H and Bhatt, Manish and Mao, Yuning and Jiang, Minqi and Parker-Holder, Jack and Foerster, Jakob and others},
  journal={Advances in Neural Information Processing Systems},
  volume={37},
  pages={69747--69786},
  year={2024}
}

@article{mouret2015illuminating,
  title={Illuminating search spaces by mapping elites},
  author={Mouret, Jean-Baptiste and Clune, Jeff},
  journal={arXiv preprint arXiv:1504.04909},
  year={2015}
}

@phdthesis{10.5555/240028,
author = {Mahfoud, Samir W.},
title = {Niching methods for genetic algorithms},
year = {1996},
publisher = {University of Illinois at Urbana-Champaign},
address = {USA},
note = {UMI Order No. GAX95-43663}
}

@inproceedings{goldberg1987genetic,
  title={Genetic algorithms with sharing for multimodal function optimization},
  author={Goldberg, David E and Richardson, Jon and others},
  booktitle={Genetic algorithms and their applications: Proceedings of the Second International Conference on Genetic Algorithms},
  volume={4149},
  pages={414--425},
  organization={Lawrence Erlbaum, Hillsdale, NJ}
}

@inproceedings{mahfoud1992crowding,
  title={Crowding and preselection revisited.},
  author={Mahfoud, Samir W and others},
  booktitle={PPSN},
  volume={2},
  pages={27--36},
  year={1992}
}

@INPROCEEDINGS{542703,
  author={Petrowski, A.},
  booktitle={Proceedings of IEEE International Conference on Evolutionary Computation}, 
  title={A clearing procedure as a niching method for genetic algorithms}, 
  year={1996},
  volume={},
  number={},
  pages={798-803},
  doi={10.1109/ICEC.1996.542703},
  ISSN={},
  month={May},}

@ARTICLE{6793380,
  author={Lehman, Joel and Stanley, Kenneth O.},
  journal={Evolutionary Computation}, 
  title={Abandoning Objectives: Evolution Through the Search for Novelty Alone}, 
  year={2011},
  volume={19},
  number={2},
  pages={189-223},
  doi={10.1162/EVCO_a_00025},
  ISSN={1063-6560},
  month={June},}

@article{guo2023connecting,
  title={Connecting large language models with evolutionary algorithms yields powerful prompt optimizers},
  author={Guo, Qingyan and Wang, Rui and Guo, Junliang and Li, Bei and Song, Kaitao and Tan, Xu and Liu, Guoqing and Bian, Jiang and Yang, Yujiu},
  journal={arXiv preprint arXiv:2309.08532},
  year={2023}
}

@article{fernando2023promptbreeder,
  title={Promptbreeder: Self-referential self-improvement via prompt evolution},
  author={Fernando, Chrisantha and Banarse, Dylan and Michalewski, Henryk and Osindero, Simon and Rockt{\"a}schel, Tim},
  journal={arXiv preprint arXiv:2309.16797},
  year={2023}
}

@article{liu2023autodan,
  title={Autodan: Generating stealthy jailbreak prompts on aligned large language models},
  author={Liu, Xiaogeng and Xu, Nan and Chen, Muhao and Xiao, Chaowei},
  journal={arXiv preprint arXiv:2310.04451},
  year={2023}
}

@article{srivastava2023no,
  title={No offense taken: Eliciting offensiveness from language models},
  author={Srivastava, Anugya and Ahuja, Rahul and Mukku, Rohith},
  journal={arXiv preprint arXiv:2310.00892},
  year={2023}
}

@article{dang2025rainbowplus,
  title={RainbowPlus: Enhancing Adversarial Prompt Generation via Evolutionary Quality-Diversity Search},
  author={Dang, Quy-Anh and Ngo, Chris and Hy, Truong-Son},
  journal={arXiv preprint arXiv:2504.15047},
  year={2025}
}

@article{10.1162/106365602320169811,
    author = {Stanley, Kenneth O. and Miikkulainen, Risto},
    title = {Evolving Neural Networks through Augmenting Topologies},
    journal = {Evolutionary Computation},
    volume = {10},
    number = {2},
    pages = {99-127},
    year = {2002},
    month = {06},
    issn = {1063-6560},
    doi = {10.1162/106365602320169811},
    url = {https://doi.org/10.1162/106365602320169811},
    eprint = {https://direct.mit.edu/evco/article-pdf/10/2/99/1493254/106365602320169811.pdf},
}

@inproceedings{cao2024defending,
  title={Defending against alignment-breaking attacks via robustly aligned llm},
  author={Cao, Bochuan and Cao, Yuanpu and Lin, Lu and Chen, Jinghui},
  booktitle={Proceedings of the 62nd Annual Meeting of the Association for Computational Linguistics (Volume 1: Long Papers)},
  pages={10542--10560},
  year={2024}
}

@inproceedings{harik1995finding,
  title={Finding multimodal solutions using restricted tournament selection.},
  author={Harik, Georges R and others},
  booktitle={ICGA},
  pages={24--31},
  year={1995}
}

@inproceedings{ando2007heuristic,
  title={Heuristic speciation for evolving neural network ensemble},
  author={Ando, Shin},
  booktitle={Proceedings of the 9th annual conference on Genetic and evolutionary computation},
  pages={1766--1773},
  year={2007}
}

@article{burlacu2023inheritance,
author = {Burlacu, Bogdan and Yang, Kaifeng and Affenzeller, Michael},
year = {2023},
month = {01},
pages = {},
title = {Population diversity and inheritance in genetic programming for symbolic regression},
volume = {23},
journal = {Natural Computing},
doi = {10.1007/s11047-022-09934-x}
}

@InProceedings{kim2009distancemeasures,
author="Kim, Kyung-Joong
and Cho, Sung-Bae",
editor="Leung, Chi Sing
and Lee, Minho
and Chan, Jonathan H.",
title="Evaluation of Distance Measures for Speciated Evolutionary Neural Networks in Pattern Classification Problems",
booktitle="Neural Information Processing",
year="2009",
publisher="Springer Berlin Heidelberg",
address="Berlin, Heidelberg",
pages="630--637",
isbn="978-3-642-10684-2"
}

@inproceedings{bhardwaj-etal-2024-language,
    title = "Language Models are {H}omer Simpson! Safety Re-Alignment of Fine-tuned Language Models through Task Arithmetic",
    author = "Bhardwaj, Rishabh  and
      Do, Duc Anh  and
      Poria, Soujanya",
    editor = "Ku, Lun-Wei  and
      Martins, Andre  and
      Srikumar, Vivek",
    booktitle = "Proceedings of the 62nd Annual Meeting of the Association for Computational Linguistics (Volume 1: Long Papers)",
    month = aug,
    year = "2024",
    address = "Bangkok, Thailand",
    publisher = "Association for Computational Linguistics",
    url = "https://aclanthology.org/2024.acl-long.762/",
    doi = "10.18653/v1/2024.acl-long.762",
    pages = "14138--14149"
}

@article{Bhardwaj2023RedTeamingLL,
  title={Red-Teaming Large Language Models using Chain of Utterances for Safety-Alignment},
  author={Rishabh Bhardwaj and Soujanya Poria},
  journal={ArXiv},
  year={2023},
  volume={abs/2308.09662},
  url={https://api.semanticscholar.org/CorpusID:261030829}
}

@misc{rit-rc-services,
author = {{Rochester Institute of Technology}},
title = {{Research Computing Services}},
year = {2019},
publisher = {{Rochester Institute of Technology}},
doi = {10.34788/0S3G-QD15},
url = {https://doi.org/10.34788/0S3G-QD15},
note = {Accessed 2026-01-23}
}

\end{document}